\pdfoutput=1

\documentclass[draftcls,12pt, onecolumn]{IEEEtran}

\date{}

\usepackage{enumitem}
\usepackage[pdftex]{graphicx}
\usepackage{tabularx}
\usepackage{amsmath} 
\usepackage{multirow,booktabs}
\usepackage{multicol}
\usepackage[justification=centering]{caption}
\usepackage{mathabx}
\usepackage{changepage}
\usepackage{url}

\newpage

\begin{document}

\begin{itemize}[align=parleft, labelsep=2.0cm]

\item[\textbf{Citation}]{D. Temel, M. Prabhushankar and G. AlRegib, "UNIQUE: Unsupervised Image Quality Estimation," in IEEE Signal Processing Letters, vol. 23, no. 10, pp. 1414-1418, Oct. 2016.
}

\item[\textbf{DOI}]{https://doi.org/10.1109/LSP.2016.2601119}

\item[\textbf{Review}]{Submitted March 29 2016, Accepted August 9 2016}

\item[\textbf{Code}]{https://ghassanalregib.com/publications/}

\item[\textbf{Bib}] {@ARTICLE\{Temel2016\_SPL,\\ 
author=\{D. Temel and M. Prabhushankar and G. AlRegib\},\\ 
journal=\{IEEE Signal Processing Letters\},\\ 
title=\{UNIQUE: Unsupervised Image Quality Estimation\}, \\ 
year=\{2016\},\\ 
volume=\{23\},\\ 
number=\{10\},\\ 
pages=\{1414-1418\},\\ 
doi=\{10.1109/LSP.2016.2601119\},\\ 
ISSN=\{1070-9908\},\\ 
month=\{Oct\},\}
} 

\item[\textbf{Copyright}]{\textcopyright 2016 IEEE. Personal use of this material is permitted. Permission from IEEE must be obtained for all other uses, in any current or future media, including reprinting/republishing this material for advertising or promotional purposes,
creating new collective works, for resale or redistribution to servers or lists, or reuse of any copyrighted component
of this work in other works. }

\item[\textbf{Contact}]{alregib@gatech.edu~~~~~~~\url{https://ghassanalregib.com/}\\dcantemel@gmail.com~~~~~~~\url{http://cantemel.com/}}
\end{itemize}
\thispagestyle{empty}
\newpage
\clearpage
\setcounter{page}{1}

\title{UNIQUE: Unsupervised Image Quality Estimation}

\author{Dogancan Temel,~\IEEEmembership{Student Member,~IEEE,}
        	  Mohit~Prabhushankar,~\IEEEmembership{Student Member,~IEEE,} and Ghassan~AlRegib,~\IEEEmembership{Senior Member,~IEEE}

\thanks{}

\thanks{The authors are with the Center for Signal and Information Processing, School
of Electrical and Computer Engineering, Georgia Institute of Technology, Atlanta,
GA, 30332 USA (e-mail: dcantemel@gmail.com; mohit.p@gatech.edu; alregib@gatech.edu) }
}

\markboth{IEEE Signal Processing Letters,~Vol.~-, No.~-, -~2016}%
{Shell \MakeLowercase{\textit{et al.}}: Bare Demo of IEEEtran.cls for Journals}

\maketitle

\begin{abstract}
In this paper, we estimate perceived image quality using sparse representations obtained from generic image databases through an unsupervised learning approach. A color space transformation, a mean subtraction, and a whitening operation are used to enhance descriptiveness of images by reducing spatial redundancy; a linear decoder is used to obtain sparse representations; and a thresholding stage is used to formulate suppression mechanisms in a visual system. A linear decoder is trained with 7 GB worth of data, which corresponds to 100,000 8x8 image patches randomly obtained from nearly 1,000 images in the ImageNet 2013 database. A patch-wise training approach is preferred to maintain local information. The proposed quality estimator UNIQUE is tested on the LIVE, the Multiply Distorted LIVE, and the TID 2013 databases and compared with thirteen quality estimators. Experimental results show that UNIQUE is generally a top performing quality estimator in terms of accuracy, consistency, linearity, and monotonic behavior.
\end{abstract}


\IEEEpeerreviewmaketitle

\section{Introduction}
\label{sec:intro}
\IEEEPARstart{I}{n} recent years, online media and social networks are dominated by a plethora of images. In these platforms, users are also content generators who contribute to the formation of big data. As data gets bigger, it becomes impossible to assess the perceived quality of images subjectively. Therefore, it is critical to automatically assess the image quality. The main challenge in quality assessment is the problem definition because it is not intuitive to define quality. If there is a distortion-free reference image, quality of an observed image can generally be approximated by quantifying pixel-wise differences between the observed stimuli and the reference. This type of quality definition is denoted as fidelity. Fidelity-based approaches have dominated image quality assessment research for a long time. However, in recent years, research community started to pay more attention to perception and its role in defining quality. It is not sufficient to study the sensory experience to model the perceptual experience of subjects because visual information is processed after acquisition. However, we can perform subjective experiments to learn mappings between pixels and perception. In the proposed work, we focus on mapping pixels to perception through data-driven learning.

Learning-based approaches are commonly used in the literature to obtain objective image quality estimators. The authors in \cite{charrier2012} propose MLIQM, which consists of detecting distortion types based on image attributes and quantifying degradation with a support-vector regression stage. A rectifier neural network-based blind quality estimator is proposed by the authors in \cite{Tang2014}, in which generic images degraded with simulated distortions are used in the pre-training stage and labeled data is used in the tuning stage. The authors in \cite{Ye2013} propose an image quality assessment approach based on filter learning. The weights are learned with a support vector regression mechanism and the filter set is learned using a stochastic gradient descent approach. Moreover, filters are also learned through k-means clustering and regression is used to obtain quality estimates. A sparse coding method is used in \cite{chang2012} to obtain abstract representations of images and Spearman correlation is used to quantify distortion.  An independent component analysis (ICA) is performed to obtain a sparse weighting matrix. ICA basis functions are obtained from the same database, in which k-fold validation is performed. The authors in \cite{Ye2012} use an unsupervised dictionary learning approach and a supervised learning approach for regression. A dictionary is learned from distorted images using k-means clustering and regression is performed to map features to scores. In \cite{Gu2014}, the authors propose an unsupervised learning approach to obtain quality-aware filters from distorted images and regression is used to obtain quality scores. In addition to dictionary-based approaches, natural scene statistics are also  used to extract descriptive features from spatial or frequency domain representations and supervised approaches are used to regress these features to quality scores \cite{Mittal2012,Moorthy2010,Saad2012}.

 All these learning-based approaches \cite{charrier2012}-\cite{Saad2012} either require subjective scores or images degraded with specific distortions. To avoid distortion specific data dependency and ground truth requirement, we propose estimating perceived quality through sparse representations learned from generic image databases. To the best of our knowledge, UNIQUE is the only quality estimator based on comparing the monotonicity of sparse representations, which does not require subjective scores or distorted images in the training phase.

\vspace{-2.5mm}
\section{Preprocessing}
\begin{figure}[htbp!]
\begin{center}
\noindent
  \includegraphics[width=0.95\linewidth]{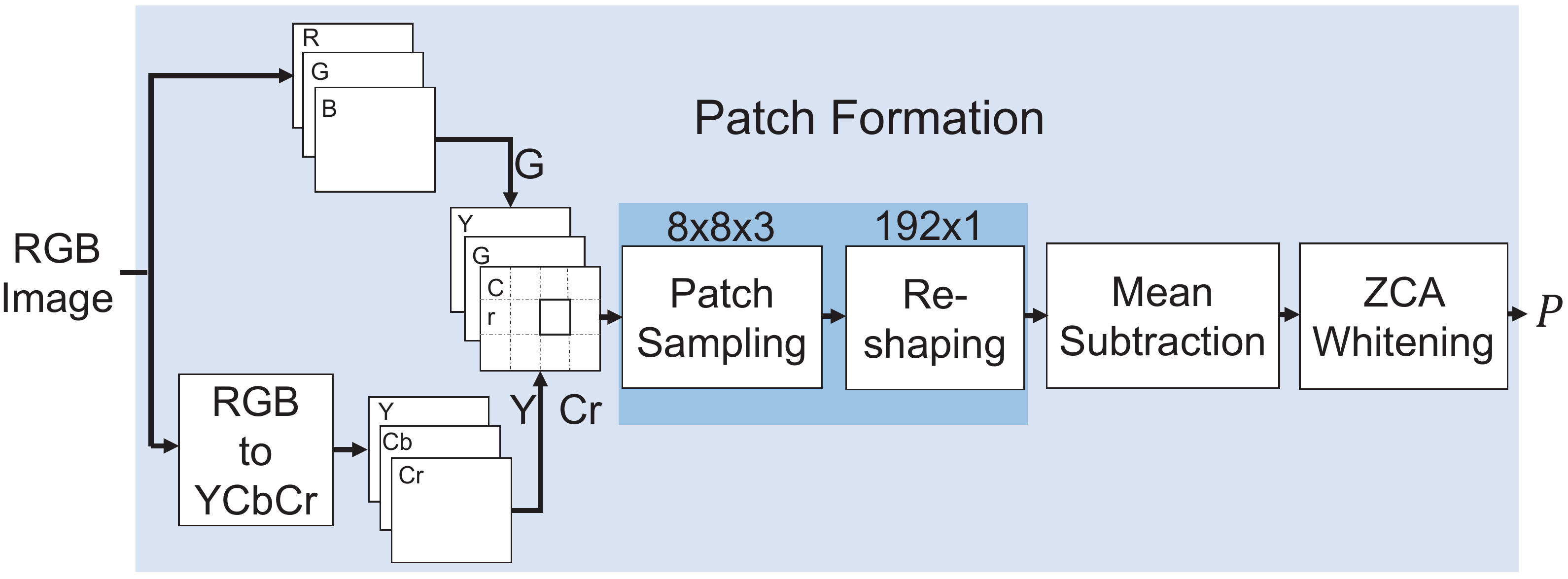}
	\vspace{-1.5mm}
    \caption{The preprocessing pipeline.}
  \label{fig:pre}
\end{center}
\vspace{-10.0mm}
\end{figure}

\vspace{-1.0mm}
Generic images are preprocessed to obtain more descriptive spatial representations. Preprocessing steps are illustrated in Fig. \ref{fig:pre}. At first, a color space selection is performed. Patches are randomly sampled over selected color channels, concatenated into a single vector, and normalized using a mean subtraction and a whitening operation. 
\vspace{-2.0mm}
\subsection{Color Space Selection}
The human visual system is more sensitive to changes in intensity compared to color as exploited in the chroma subsampling \cite{lamb2001}. However, color channels still contain  information that is not conveyed by intensity. An intuitive way to introduce color information is to directly use RGB channels. However, there is a high correlation between color channels. The G channel has most of the information already contained in the R and the B channels \cite{tkalcic2003colour} so we use the G channel. We transform RGB images into YCbCr images and use the Y channel, which includes structural information \cite{wang2004image}. Moreover, we also use the Cr channel based on empirical studies.
\vspace{-4.5mm}
\subsection{Patch-based Sampling and Reshaping}
\vspace{-1.0mm}
An $8x8$ patch is randomly sampled from an input image and converted into a $1$-$D$ vector of length $64$. Three channels are concatenated into a $1$-$D$ vector of length $192$. 
\vspace{-4.5mm}
\subsection{Mean Subtraction and ZCA Whitening}
For each location in these $1$-$D$ representations, we compute a mean value over all patches and perform mean subtraction. Then, we perform ZCA whitening patch-wise to decorrelate spatial representations as explained in \cite{chang2012}. 
\vspace{-2.0mm}

\section{Unsupervised Image Quality Estimation}

\vspace{-1.0mm}

\subsection{Training set}
\vspace{-1.0mm}
Quality estimators are usually trained using image quality databases or simulated distortions, which can limit quality assessment capability to specific distortions. Therefore, to avoid overfitting, we use the ImageNet database, whose images contain a queried object along with other objects, multiple instances, occlusion or text \cite{ImageNet_VSS09}. In our training phase, we randomly select around $1,000$ images and  extract $100$ patches from each image, which leads to a total of $100,000$ patches.
\vspace{-3.5mm}

\subsection{Sparse Representation}
\vspace{-1.0mm}
The authors in \cite{olshausen1997sparse} claim that sparse coding, with an overcomplete basis set, operates similar to encoding mechanisms of visual representations in the V1 cortex and response characteristics of simple V1 cells can be simulated by learning weight parameters over patches. In the proposed work, we use a linear decoder architecture to obtain sparse representations.

\begin{figure}[htbp!]
	\begin{center}
		\noindent
		\includegraphics[width=0.6\linewidth]{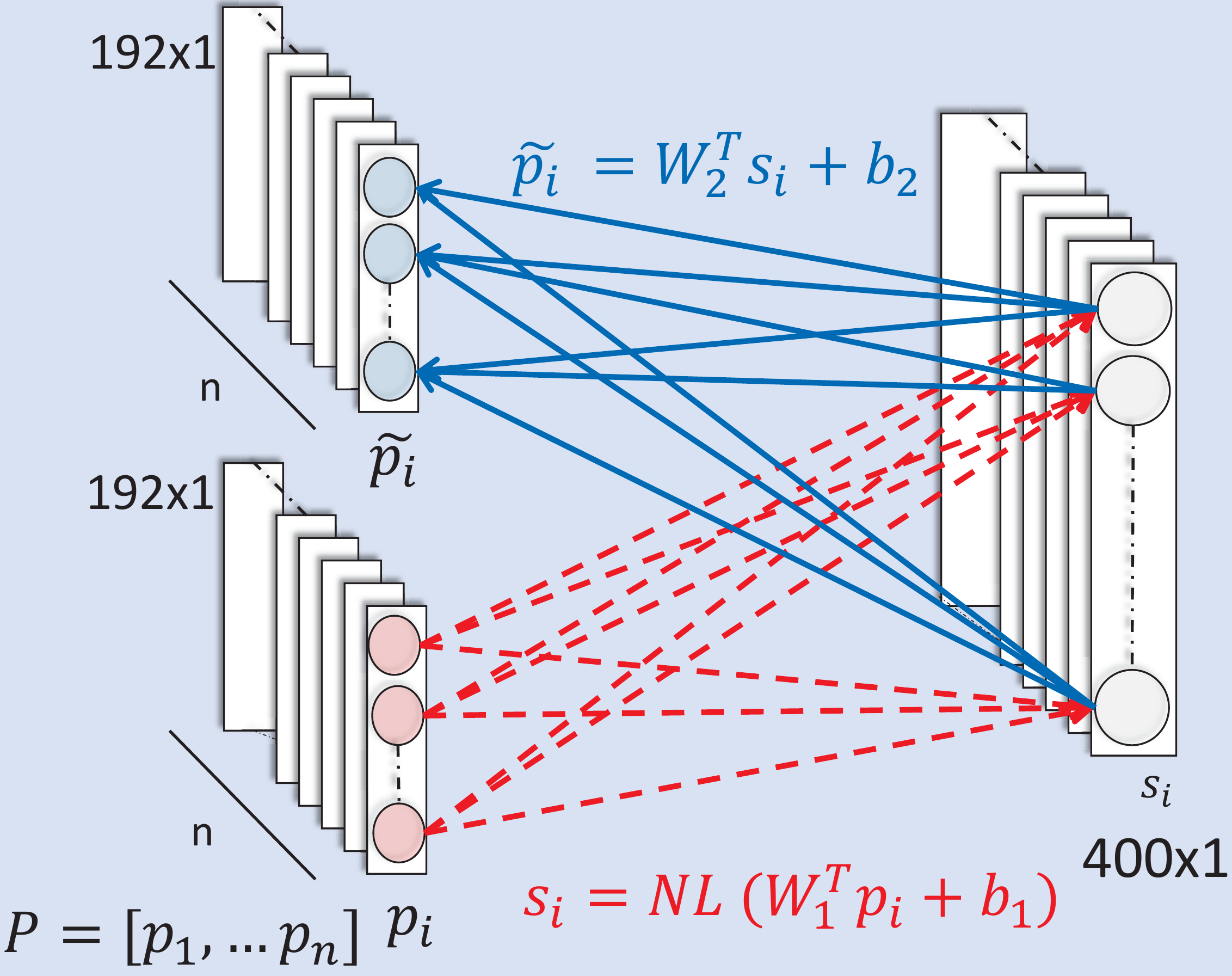}
		\vspace{-1.5mm}
	    \caption{Linear decoder architecture, in which NL is a non-linear sigmoid layer.}
		\vspace{-7.0mm}
		\label{fig:decoder}
	\end{center}
\end{figure}

\subsection{Unsupervised Learning}
\begin{figure}[htbp!]\label{Visualize}
	\begin{center}
		\noindent
		\includegraphics[width=0.6\linewidth, trim= 4mm 4mm 4mm 4mm]{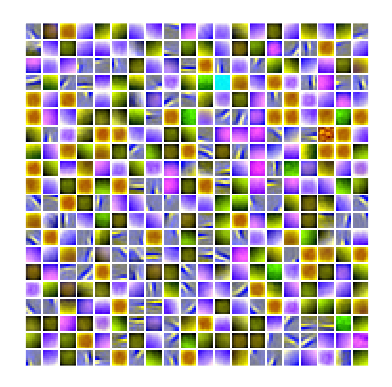}
		\vspace{-1.0mm}
		\caption{Visualization of learned features.}
		\vspace{-1.0mm}
		\label{fig:vis}
	\end{center}
\vspace{-8.0mm}
\end{figure}

A linear decoder is an unsupervised learning architecture used to represent input data in different dimensions. The output of a linear decoder is a reconstructed version of the input and a backpropagation operation is performed to reduce the reconstruction error by adjusting weights and bias. We set the target dimension higher than the input and add a sparsity constraint. The sparse representation $s$ is obtained using the whitened map $P$ as 
\vspace{-3.5mm}
\begin{equation}\label{AutoSparse}
s = W_1^T P + b_1,
\vspace{-1.5mm}
\end{equation}
where $W_1$ and $b_1$ refer to a weight matrix and a bias vector, respectively. $s$ is passed through a non-linear sigmoid activation function.  The objective function $J(W,b)$ is minimized using a backpropogation operation, in which $W$ includes reconstruction weights $W_2$ and bias $b_2$ in addition to $W_1$ and $b_1$. Adding a sparsity penalty term weighted by $\beta$, the objective function is expressed as
\vspace{-3.0mm}
\begin{equation} \label{decoder_cost}
	J(W,b) = \lVert (W_2^Ts + b_2) - P \rVert_2^2 + \beta \sum_{j=1}^{N} {\rm KL}(\rho || \hat\rho_j) + \lambda\lVert W \rVert_2^2,
\vspace{-0.5mm}
\end{equation}
where the first term is the reconstruction error, the second term is the sparsity penalty, and the third term is the weight decay. Note that $s$ is a function of $W_1$ and $b_1$ as in (\ref{AutoSparse}) and minimization is performed over weights and bias in both encoding and decoding stages. The weight decay term corresponds to regularization, which limits weights and prevent overfitting to only particular input units. Minimization is carried out using limited memory BFGS (L-BFGS) algorithm. To account for non-smoothness, we use KL-divergence, which can be expressed as 
\vspace{-2.0mm}
\begin{equation} \label{kl_divergence}
\sum_{j=1}^{N} {\rm KL}(\rho || \hat\rho_j) = \sum_{j=1}^{N} \rho \log \frac{\rho}{\hat\rho_j} + (1-\rho) \log \frac{1-\rho}{1-\hat\rho_j},
\vspace{-1.0mm}
\end{equation}
where N is the number of hidden units, $\rho$ is the target average activation (set to $0.035$), and $\hat\rho$ is the actual average activation formulated as
\vspace{-3.5mm}
\begin{equation}
\hat\rho = \frac{1}{M}\sum_{i=1}^{M} s_i.  
\vspace{-1.0mm}
\end{equation}
where $M$ is the number of training examples in one forward pass. This type of objective function definition not only leads to smoothening but also preserves sparsity in the hidden units \cite{bradley2008differential}.
$\beta$ is set to $5$ based on empirical studies
to control the weight of the sparsity penalty term. The linear decoder architecture is shown in Fig. \ref{fig:decoder}. An image patch of size $192$x$1$ is mapped to a sparse representation of size $400$x$1$. The nodes in $s_i$ represent hidden layer units. In Fig. \ref{fig:vis}, each of the squares depict learned weights, which can be used to infer the patches that maximally activate hidden units. A weighted sum of these hidden units is used to approximate natural images.
\subsection{Image Quality Estimation}
\begin{figure}[htbp!]
	\begin{center}
		\noindent
		\includegraphics[width=0.8\linewidth]{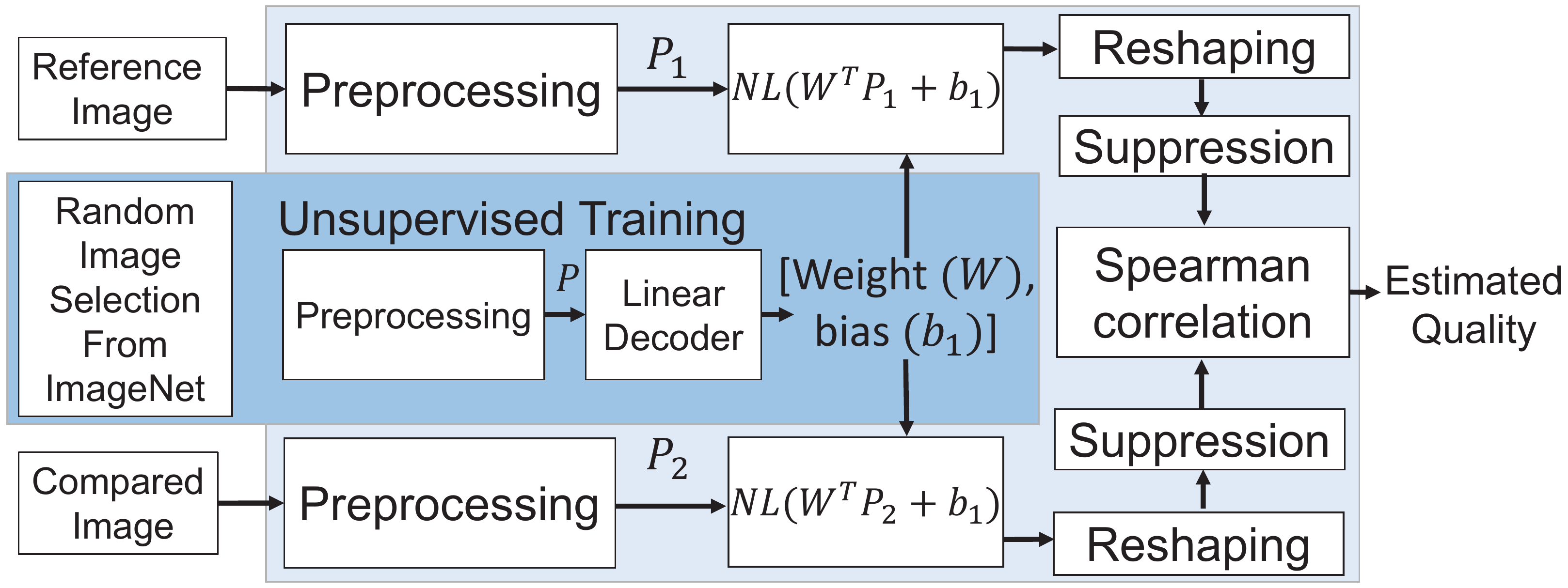}
		\caption{The proposed image quality estimation pipeline.}
		\label{fig:main}
	\end{center}
	\vspace{-6.0mm}
\end{figure}

\vspace{-0.5mm}
The proposed quality estimation pipeline is given in Fig. \ref{fig:main}.
A linear decoder is trained in an unsupervised fashion by feeding preprocessed patches. The learned weights and the bias are used along with a non-linear mapping to transform preprocessed non-overlapping patches of the reference and the compared images into sparse representations. These representations are reshaped into vectors. If an entity in these vectors is significantly less than the average activation value, a zero is assigned to mimic suppression mechanisms in a visual system. Spearman rank order correlation coefficient is used to compare two reshaped vectors and we use $10^{th}$ power of the correlation coefficient to utilize full quality estimate range.
	\vspace{-1.0mm}

Contrary to sparse coding approaches, we do not approximate natural image patches using overcomplete basis. We treat learned weights as filters and use inner product to calculate the response of each patch to every filter. In effect, we project input patches on the filters and obtain sparse representations based on their responses, which are hidden unit activations. These sparse representations are suppressed and compared to obtain image quality scores. Sparse coding-based methods generally perform optimization even during the testing stage, which is not performed in the proposed method. 
%

\section{Validation}
\vspace{-3.0mm}
\subsection{Databases}
In the comparison of the quality estimators, we use the LIVE \cite{live2006}, the Multiply Distorted LIVE (MULTI) \cite{multi2012}, and the TID 2013 (TID13) \cite{tid13} databases. Distortion types in these databases can be classified into seven groups. The compression group consists of JPEG, JPEG2000, and lossy compression of noisy images. The noise group includes additive Gaussian noise, additive noise, spatially correlated noise, masked noise, high frequency noise, impulse noise, quantization noise, image denoising, multiplicative Gaussian noise, comfort noise, and lossy compression of noisy images. The communication group includes JPEG and JPEG2000 transmission errors. The blur group consists of Gaussian blur and sparse sampling and reconstruction. The color group contains change of color saturation, image color quantization with dither and chromatic aberrations. The global category includes intensity shift and contrast change. The local group contains non-eccentricity pattern noise and local block-wise distortions of different intensity. 


\subsection{Performance Metrics and Compared Quality Estimators}
\label{subsec:val_metrics}
The performance of the quality estimators are validated using root mean square error (accuracy), outlier ratio (consistency), Pearson correlation (linearity), and Spearman correlation (monotonic behavior). In the outlier ratio calculation, we measure the outliers that are more than two standard deviations away from the average subjective scores and outlier ratios are only reported for MULTI and TID13 because standard deviations of subjective scores are not reported in the LIVE database. Statistical significance between correlation coefficients is measured with the formulations suggested in ITU-T Rec. P.1401. \cite{ITU-T}. We use the regression formulation suggested in \cite{live2006} because performance metrics are sensitive to the range and the distribution of the scores. Moreover, we measure the difference between the normalized histograms of subjective scores and the regressed quality estimates through common histogram difference metrics including Earth Mover's Distance (EMD), Kullback-Leibler (KL) divergence,  Jensen-Shannon (JS) divergence, histogram intersection (HI), and L2 norm. The lower the difference, the better the estimation performance. In the performance comparison, we use full-reference methods based on fidelity, perceptually-extended fidelity, structural similarity, spectral similarity, feature similarity, and perceptual similarity, and no-reference methods based on natural scene statistics.
\vspace{-4.0mm}

\begin{center}
\begin{table*}[htbp!]
\scriptsize
\centering
\caption{Performance of image quality estimators.}
\label{tab_results_databases}
\vspace{-2.0mm}
\begin{tabular}{p{0.9cm}|p{0.63cm}|p{0.63cm}|p{0.63cm}|p{0.63cm}|p{0.63cm}|p{0.63cm}|p{0.63cm}|p{0.60cm}|p{0.75cm}|p{0.75cm}|p{0.63cm}|p{0.63cm}|p{0.63cm}|p{0.9cm}}
\hline

\multirow{3}{*}{{\bf Methods}}                & \multirow{2}{*}{{\bf PSNR}}  & {\bf PSNR } &{\bf PSNR } &\multirow{2}{*}{{\bf SSIM }} &{\bf MS } &{\bf CW }&{\bf IW } &{\bf SR }   &\multirow{2}{*}{{\bf FSIMc }} &\multirow{2}{*}{{\bf PerSIM }}&{\bf BRIS } & \multirow{2}{*}{{\bf BIQI}} &{\bf BLII } &\multirow{2}{*}{{\bf UNIQUE }}  \\
& &\textbf{HA} &\textbf{HMA} & &{\bf SSIM } &{\bf SSIM } &{\bf SSIM } &{\bf SIM }  &&&\bf QUE & &\bf NDS2 & \\ & &\cite{Ponomarenko2011} & \cite{Ponomarenko2011} &\cite{wang2004image} &\cite{Wang2003} &\cite{Sampat2009}&\cite{Wang2011}&\cite{Zhang12} &\cite{Zhang2011}&\cite{Temel2015}&\cite{Mittal2012} &\cite{Moorthy2010} &\cite{Saad2012}&
\\ \hline

                & \multicolumn{14}{c}{\textbf{Outlier Ratio}}                                                                                                                                                                        \\ \hline

\textbf{MULTI}  
       &0.008   & 0.013  &  0.008   & 0.015  & 0.013    &0.093  & 0.013        &\bf  0   & 0.015   & 0.004    &0.068  & 0.024 &0.077 & \bf  0
 \\ 
\textbf{TID13}   
  &0.725   & \bf 0.615   & 0.670  &  0.733  &  0.691 &    0.855 &    0.700 & \bf   0.632 &    0.727 &    0.655  &0.851 &0.855 &0.851  &   0.641
\\ \hline

                & \multicolumn{14}{c}{\textbf{Root Mean Square Error}}                                                                                                                                                                        \\ \hline
\textbf{LIVE}  &8.61   & 6.93   & \bf 6.58  &  7.52  &  7.44 &  11.2 &   7.11 &   7.54 &   7.20 &   6.80  &8.57  &10.8 & 9.04  &\bf 6.76 \\ 
\textbf{MULTI}    &12.7  & 11.3 &  10.7  & 11.0   &11.2 &  18.8 &  10.0 &  \bf  8.68 &  10.7 &   9.89  &15.0 &12.7 &17.4 &\bf 9.25
 \\ 
\textbf{TID13}     &0.87   & 0.65  &  0.69 &   0.76 &   0.69  &  1.20  &  0.68 &  \bf 0.61  &  0.68 &   0.64 &1.10 & 1.10 &1.09 &   \bf 0.61
            \\ \hline              
               
               & \multicolumn{14}{c}{\textbf{Pearson Correlation Coefficient}}                                                                                                                                                                        \\ \hline

\multirow{2}{*}{{\bf LIVE}}                 & 0.928             &0.953                  &\bf 0.958                  &0.945           &0.946                   &0.872                  & 0.951               &0.945                                &0.950             & 0.955       &0.928    &  0.883  &0.920   & \bf 0.956                   \\
 &  (-1)             &               (0) &\bf  (0)                  & (-1)               & (-1)                  &  (-1)                 &   (0)                & (-1) & (0)               &  (0)        & (-1)    & (-1)   &(-1) & \bf Ref           \\

\multirow{2}{*}{{\bf MULTI}}               &0.739                &0.801                   &0.821                   &              0.812 & 0.802               &0.379                 &0.847                &\bf 0.888                                & 0.821               &  0.852           &0.605   &0.738 &0.389 &\bf 0.872                      \\
 &(-1)               &(-1)                  & (-1)                   &              (-1) & (-1)                 &(-1)                  &(0)                 &\bf                                (0) & (-1)               &  (0)        & (-1)  &(-1)  & (-1)  &\bf Ref              \\

\multirow{2}{*}{{\bf TID13}}       &0.705                & 0.850                  &0.827                    &0.788         &0.830                 &0.227                  &0.831                 &\bf 0.866            &0.832                              & 0.854       &0.460 &0.448   &0.473     &\bf 0.868                            \\&(-1)               & (-1)                 & (-1)                   & (-1)        & (-1)                &(-1)                 & (-1)                  &\bf                (0) & (-1)                              &  (-1)          &(-1)   &(-1) &(-1)  &\bf Ref              \\ \hline

\textbf{}      & \multicolumn{14}{c}{\textbf{Spearman Correlation Coefficient}}                                                                                                                                                                        \\ \hline

\multirow{2}{*}{{\bf LIVE}}             &0.909                &0.937                 &0.944                 &              0.949  & 0.951                 & 0.902                  & \bf 0.960                 & 0.955                        & \bf 0.959              & 0.950      &   0.939    &0.897 &0.922    &0.952                 \\ &(-1)               &(-1)                 & (0)                  &              (0) &  (0)                 &  (-1)                 & \bf  (1)                 & (0)                         & \bf  (0)            & (0)       &   (-1)   &(-1)  &(-1) & Ref      \\

\multirow{2}{*}{{\bf MULTI}}            &0.677              & 0.714                 & 0.743                  &              0.860  &0.836                  & 0.630                &\bf 0.883                  &                \bf 0.866                            & \bf 0.866               &  0.818      & 0.598  &0.610 &0.386 &\bf 0.866   \\  &(-1)              & (-1)                 & (-1)                  & (0)  & (0)                  & (-1)                &\bf  (0)                  &                \bf  (0)                            & \bf  (0)               &    (-1)    &  (-1)  &(-1) &(-1)  &\bf Ref        \\

\multirow{2}{*}{{\bf TID13}}          & 0.700              & 0.847                 & 0.817                  &0.741        &0.785                & 0.562               & 0.777              &0.807             &0.851                             &\bf 0.853      & 0.414   &0.393   &0.396  &\bf 0.860  
   \\  & (-1)              & (0)                 &  (-1)                  & (-1)        & (-1)                & (-1)               &  (-1)              &  (-1)             & (0)                             &\bf (0)           &  (-1) &(-1) &(-1) &\bf Ref                        
     \\ \hline

\end{tabular}
\end{table*}
\end{center}

\begin{center}

\begin{table*}[htbp!]
  \centering
      \scriptsize
      \vspace{-2.0mm}
        \caption{Distributional differences between subjective scores and objective quality estimates.}
        \vspace{-2.0mm}

    \begin{tabular}{c||c|c|c|c|c||c|c|c|c|c||c|c|c|c|c}   \hline
    \multirow{2}[4]{*}{\textbf{Metric}} 
    
    & \multicolumn{5}{c||}{\textbf{Difference-LIVE}} & \multicolumn{5}{c||}{\textbf{Difference-MULTI}} & \multicolumn{5}{c}{\textbf{Difference-TID13}} \\ \cline{2-16}     
    
          & \textbf{EMD}& \textbf{KL}& \textbf{JS}& \textbf{HI}& \textbf{L2}& \textbf{EMD}& \textbf{KL}& \textbf{JS}& \textbf{HI}& \textbf{L2}  & \textbf{EMD}& \textbf{KL}& \textbf{JS}& \textbf{HI}& \textbf{L2}    
          \\ \hline

	\textbf{IW-SSIM \cite{Wang2011}}  &  0.29  &  0.32  &  0.07 &   0.29  &  0.07 &   0.42   & 0.47 &  0.09  &  0.42  &  0.11 &   0.50 &   1.67 &   0.19 &   0.50  &  0.18
  \\ 

	\textbf{SR-SIM \cite{Zhang12}}   &  0.32  &  0.38  &  0.08 &   0.32  &  0.08 &    0.40  &  0.42  &  0.09  &  0.40  & 0.10 &   0.50 &   1.62 &   0.19 &   0.50  &  0.17
 \\ 
	
		\textbf{FSIMc \cite{Zhang2011}}  &  0.27  &  0.30  &  0.06  &  0.27   & 0.07 &   0.45  &  0.51 &  0.11  &  0.45  &  0.11 &   0.68  &  2.54 &   0.30  &  0.68 &   0.23
 \\

				\textbf{UNIQUE}  &  \bf 0.23  & \bf 0.25 &  \bf 0.05  &   \bf 0.23  & \bf 0.06 &  \bf 0.30 &  \bf 0.23  & \bf 0.05 &  \bf 0.30  &  \bf 0.08 &  \bf  0.40 &  \bf 0.93 & \bf  0.13 &  \bf 0.40  & \bf 0.11
\\ \hline
	
    \end{tabular}%
  \label{tab:hist_dist}
\vspace{-4.0mm}

\end{table*}

\end{center}

\vspace{-12.0mm}

\subsection{Results}
\label{subsec:val_results}
The performance of fourteen quality estimators over three databases is summarized in Table \ref{tab_results_databases}. We use \texttt{fitnlm} function in MATLAB \textsuperscript{\textregistered} to obtain regression curves. We set the initialization coefficients same for each quality estimator  as $[0.0,0.1,0.0,0.0,0.0]$, from  which a non linear model starts its search for optimal coefficients. Regression operation leads to a parabolic behavior for PSNR, PSNR-HA, PSNR-HMA, BRISQUE, and BIQI in the TID13 database, and for BRISQUE and BLIINDS2 in the MULTI database. Therefore, reported performances of existing methods can vary from the literature because of the differences in regression curves and initialization coefficients. In each category, we highlight the results of two best performing methods with a bold typeset. We highlight more than two methods when they lead to equivalent performances. Statistical significance with respect to correlation values of \texttt{UNIQUE} are reported between parentheses under correlation values of other methods. A $0$ corresponds to statistically similar performance, $-1$ means compared method is statistically inferior, and $1$ means compared method is statistically superior. \texttt{UNIQUE} is among the best performing methods in all categories other than Spearman correlation in the LIVE database and outlier ratio in the TID13 database. PSNR-HA, PSNR-HMA, IW-SSIM, SR-SIM, FSIMc, and PerSIM are among the best performing methods as well. SR-SIM is also consistently among the best performing methods but \texttt{UNIQUE} statistically outperforms SR-SIM in two out of six correlation categories. There is only one correlation category, in which \texttt{UNIQUE} is statistically outperformed by IW-SSIM. However, in two out of the remaining five categories, \texttt{UNIQUE} statistically outperforms IW-SSIM. Even though \texttt{UNIQUE} is a full-reference method, it is compared against state-of-the-art no-reference image quality assessment (NR-IQA) methods  BRISQUE \cite{Mittal2012}, BIQI \cite{Moorthy2010}, and BLIINDS2 \cite{Saad2012}. We use the NR-IQA methods trained on the entire LIVE database, which are provided in \cite{LIVEWeb}. In the LIVE database, even NR-IQA methods are trained and tested on the same image set, \texttt{UNIQUE}  still statistically outperforms these NR-IQA methods. 

\vspace{-4.0mm}

\begin{center}

\begin{figure}[htbp!]
\begin{minipage}[b]{0.28\linewidth}
  \centering
\includegraphics[width=0.8\linewidth, trim= 20mm 75mm 20mm 65mm]{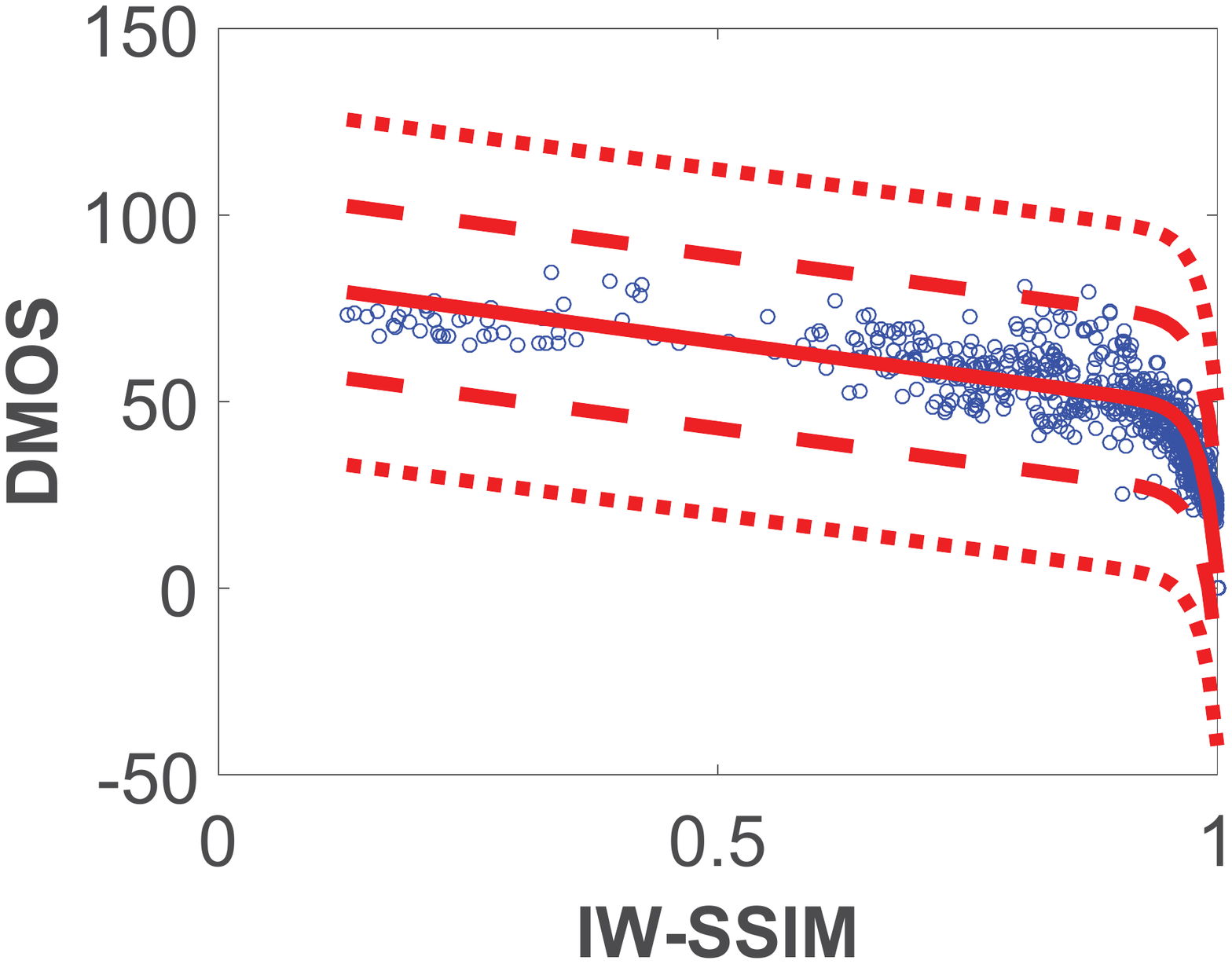}
  \vspace{0.03cm}
  \centerline{\footnotesize{(a)LIVE-IW-SSIM}}
      \vspace{-0.45cm}
\end{minipage}
 \vspace{0.2cm}
\hfill
\begin{minipage}[b]{0.28\linewidth}
  \centering
\includegraphics[width=0.8\linewidth, trim= 20mm 75mm 20mm 65mm]{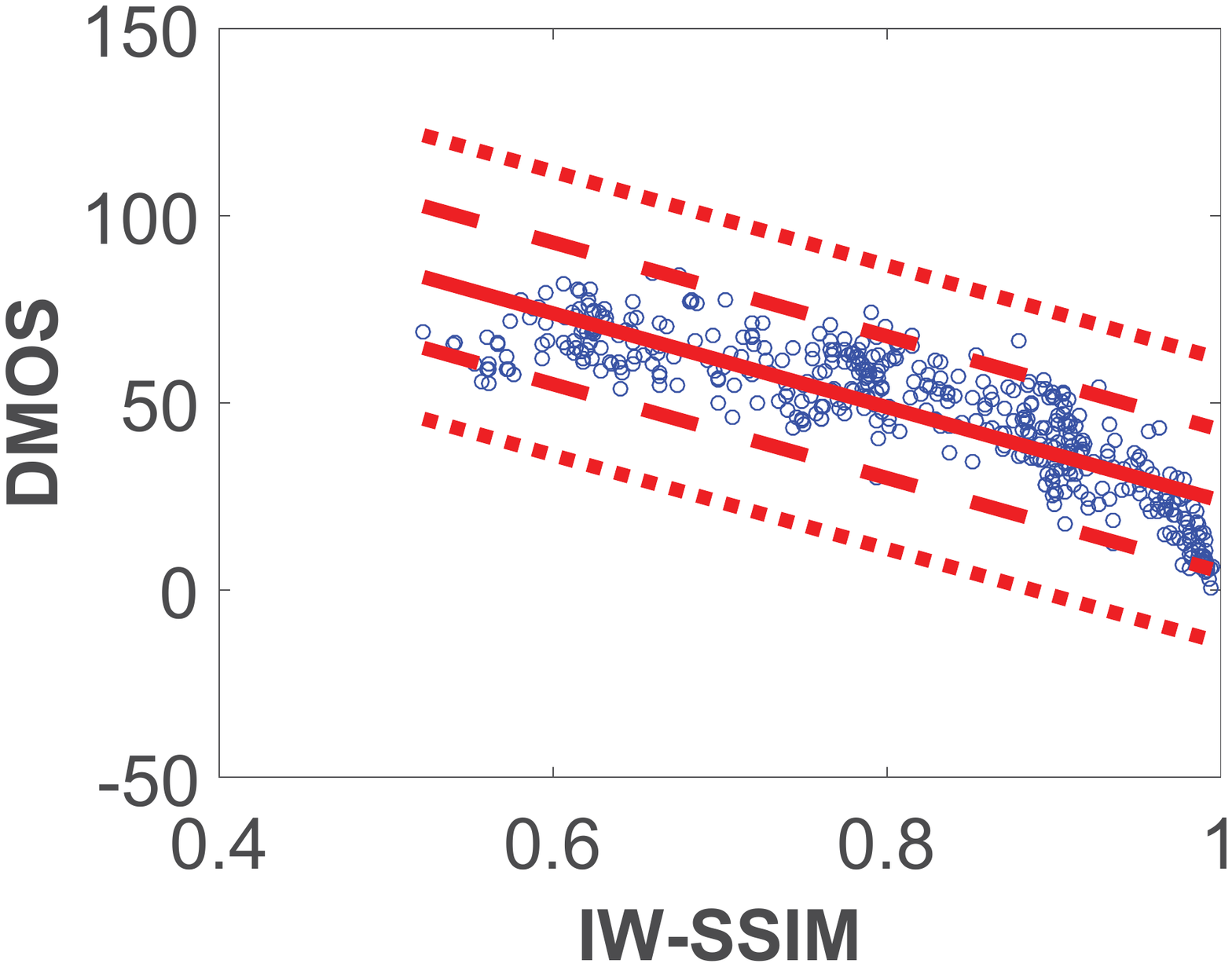}
  \vspace{0.03 cm}
  \centerline{\footnotesize{(b) MULTI-IW-SSIM   } }
      \vspace{-0.45cm}
\end{minipage}
 \vspace{0.2cm}
\hfill
\begin{minipage}[b]{0.28\linewidth}
  \centering
\includegraphics[width=0.8\linewidth, trim= 20mm 75mm 20mm 65mm]{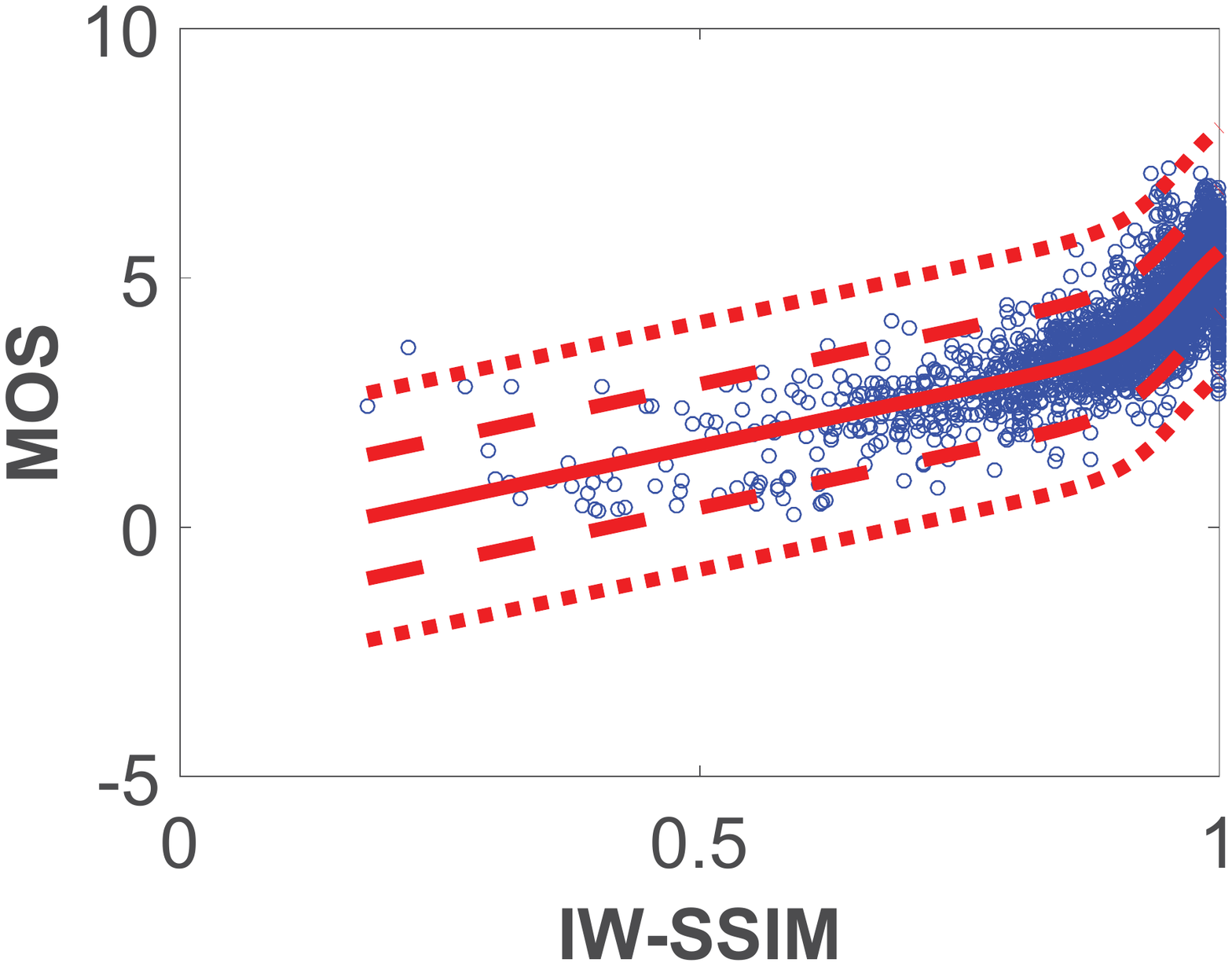}
  \vspace{0.03cm}
  \centerline{\footnotesize{(c) TID-IW-SSIM }}
      \vspace{-0.45cm}
\end{minipage}

\begin{minipage}[b]{0.28\linewidth}
  \centering
\includegraphics[width=0.8\linewidth, trim= 20mm 75mm 20mm 65mm]{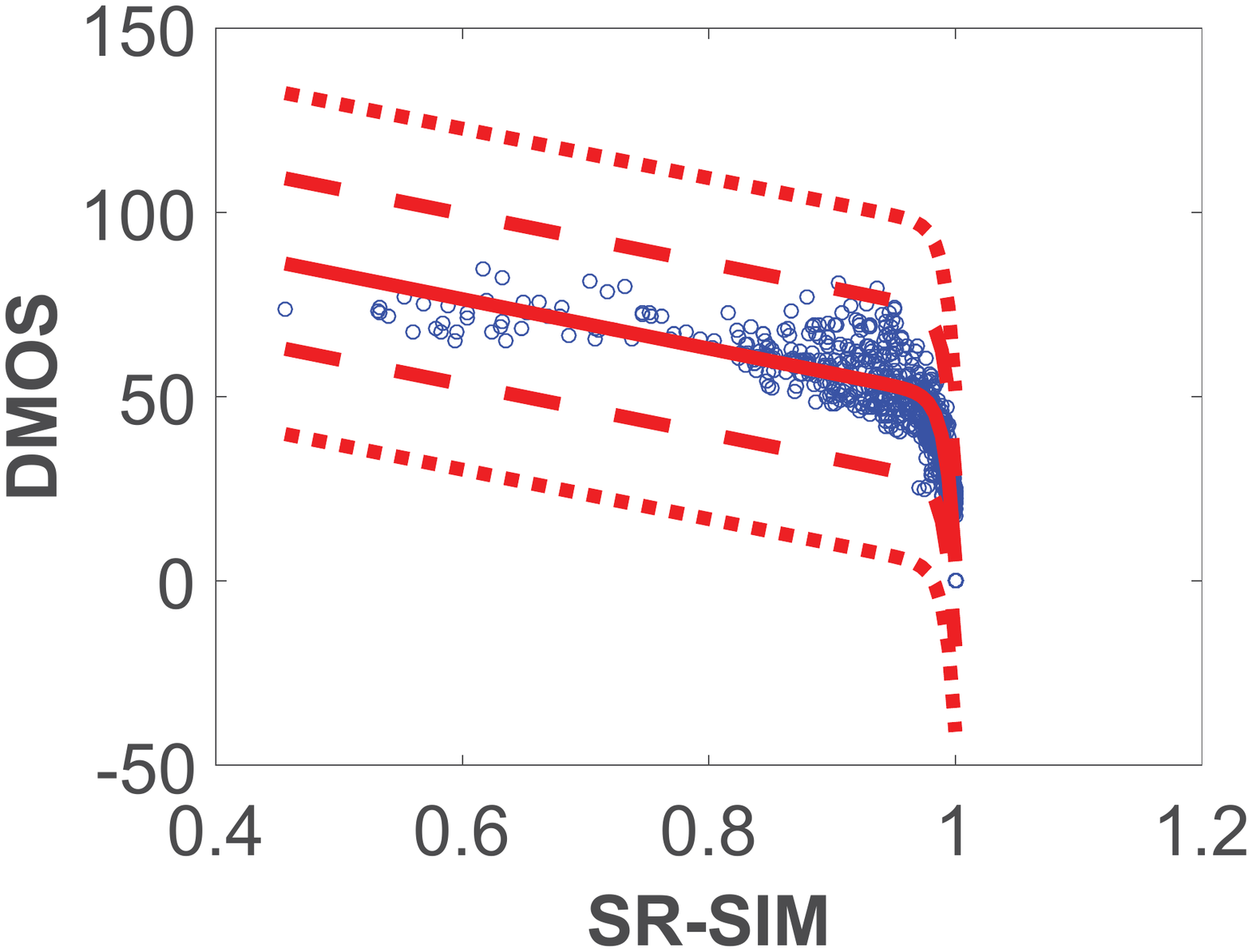}
  \vspace{0.03cm}
  \centerline{\footnotesize{(d)LIVE-SR-SIM}}
      \vspace{-0.45cm}
\end{minipage}
 \vspace{0.2cm}
\hfill
\begin{minipage}[b]{0.28\linewidth}
  \centering
\includegraphics[width=0.8\linewidth, trim= 20mm 75mm 20mm 65mm]{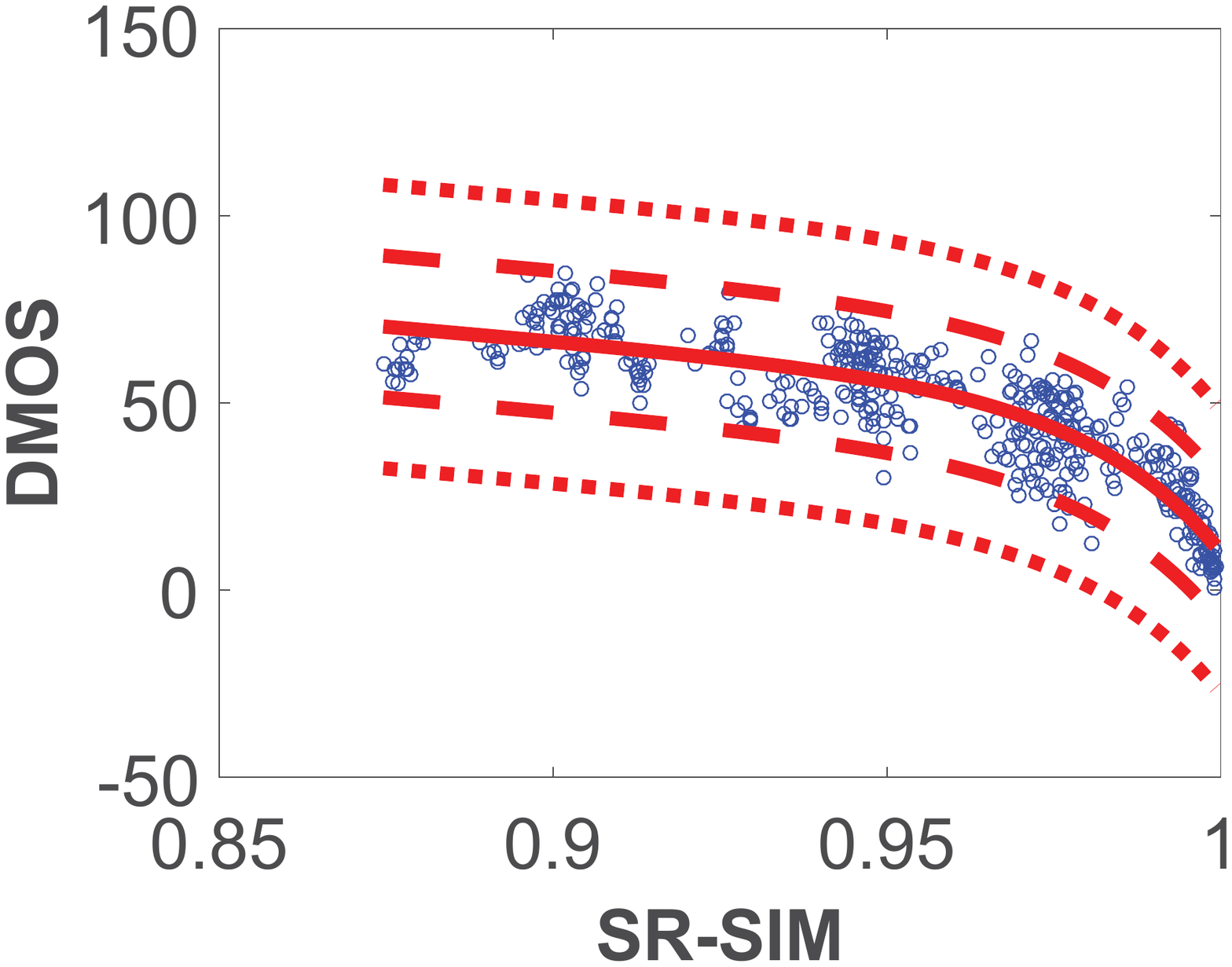}
  \vspace{0.03 cm}
  \centerline{\footnotesize{(e) MULTI-SR-SIM   } }
      \vspace{-0.45cm}
\end{minipage}
 \vspace{0.2cm}
\hfill
\begin{minipage}[b]{0.28\linewidth}
  \centering
\includegraphics[width=0.8\linewidth, trim= 20mm 75mm 20mm 65mm]{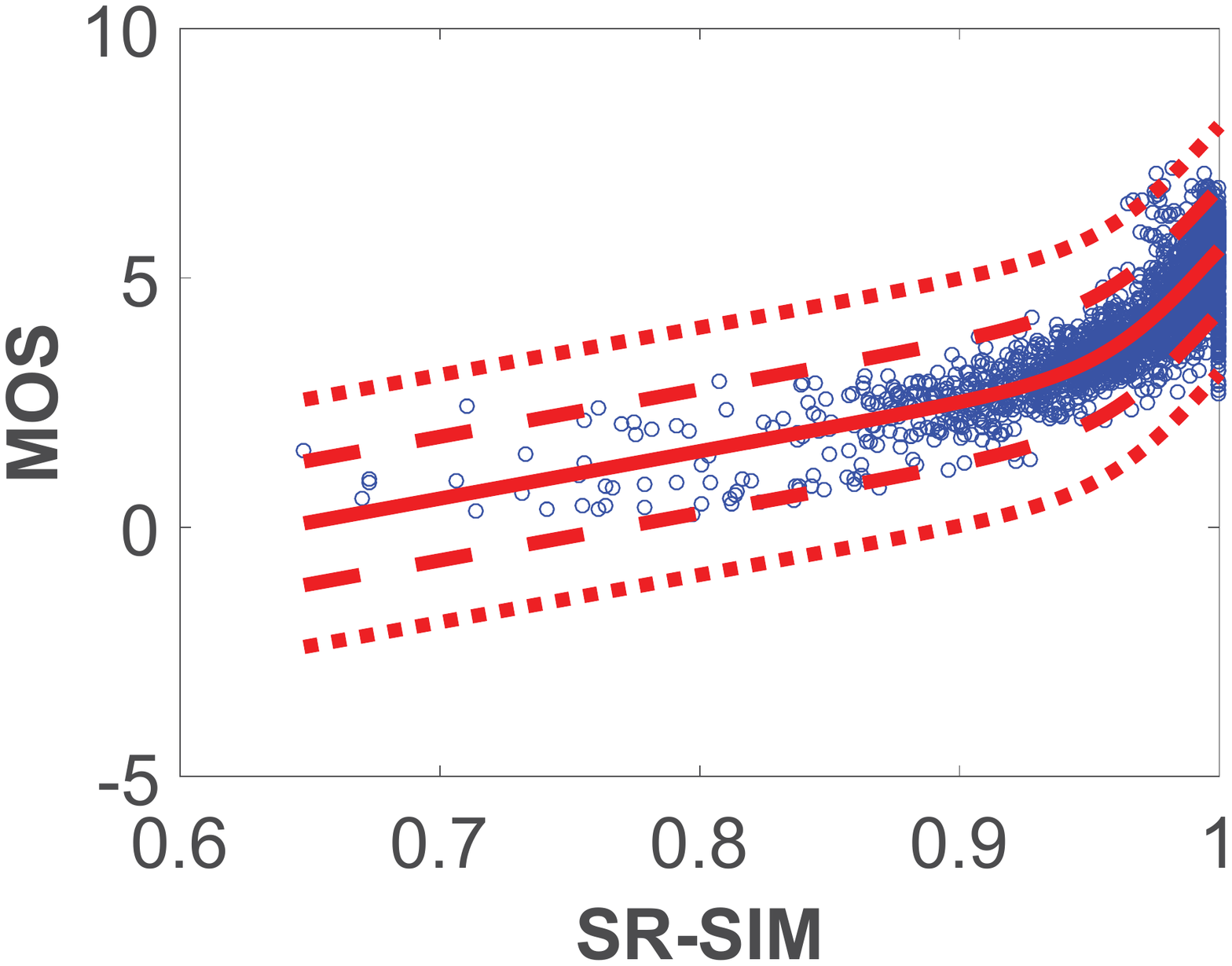}
  \vspace{0.03cm}
  \centerline{\footnotesize{(f) TID-SR-SIM }}
      \vspace{-0.45cm}
\end{minipage}

\begin{minipage}[b]{0.28\linewidth}
  \centering
\includegraphics[width=0.8\linewidth, trim= 20mm 75mm 20mm 65mm]{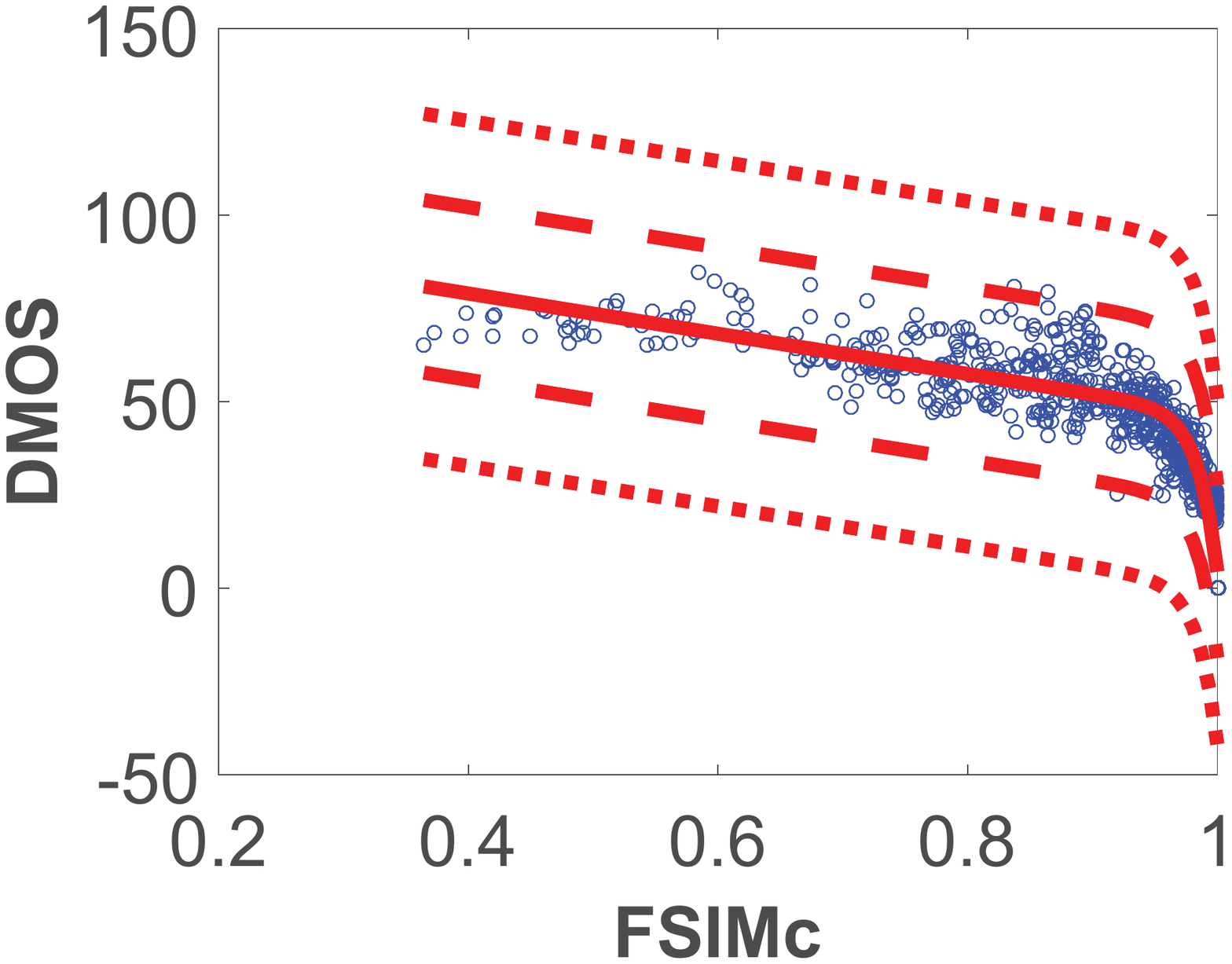}
  \vspace{0.03cm}
  \centerline{\footnotesize{(g)LIVE-FSIMc}}
      \vspace{-0.40cm}
\end{minipage}
 \vspace{0.2cm}
\hfill
\begin{minipage}[b]{0.28\linewidth}
  \centering
\includegraphics[width=0.8\linewidth, trim= 20mm 75mm 20mm 65mm]{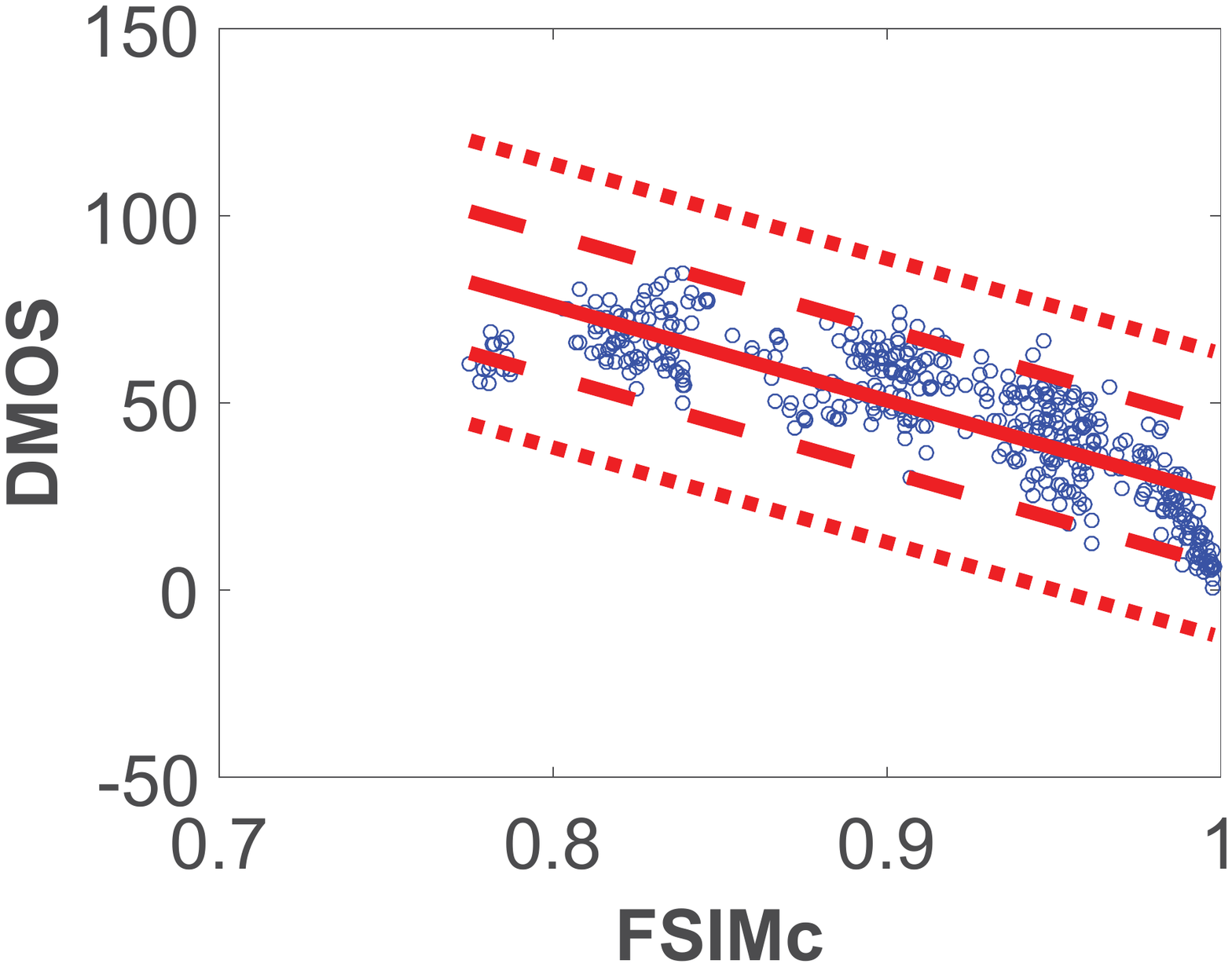}
  \vspace{0.03 cm}
  \centerline{\footnotesize{(h) MULTI-FSIMc   } }
      \vspace{-0.40cm}
\end{minipage}
 \vspace{0.2cm}
\hfill
\begin{minipage}[b]{0.28\linewidth}
  \centering
\includegraphics[width=0.8\linewidth, trim= 20mm 75mm 20mm 65mm]{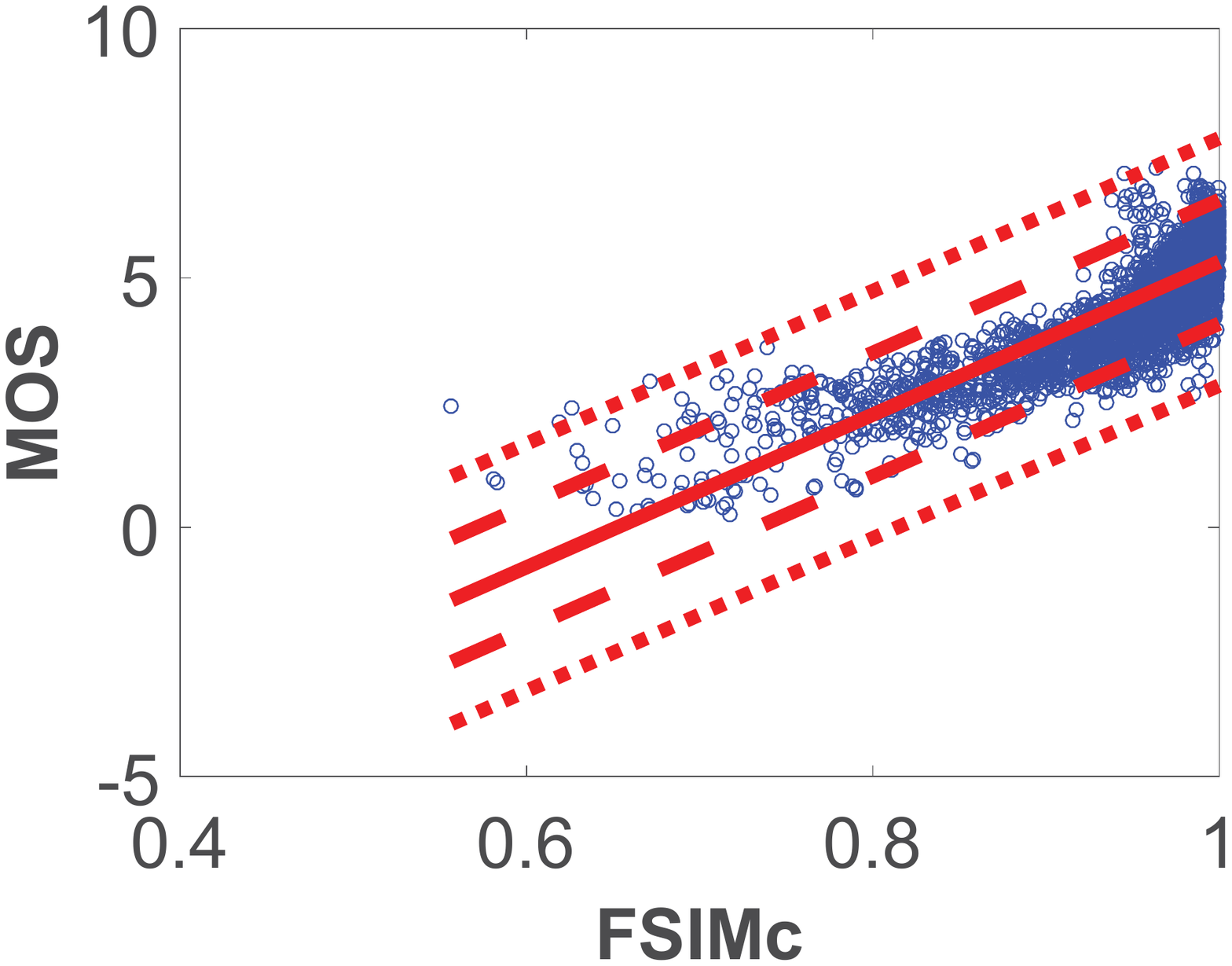}
  \vspace{0.03cm}
  \centerline{\footnotesize{(i) TID-FSIMc }}
      \vspace{-0.40cm}
\end{minipage}

\begin{minipage}[b]{0.28\linewidth}
  \centering
\includegraphics[width=0.8\linewidth, trim= 20mm 75mm 20mm 65mm]{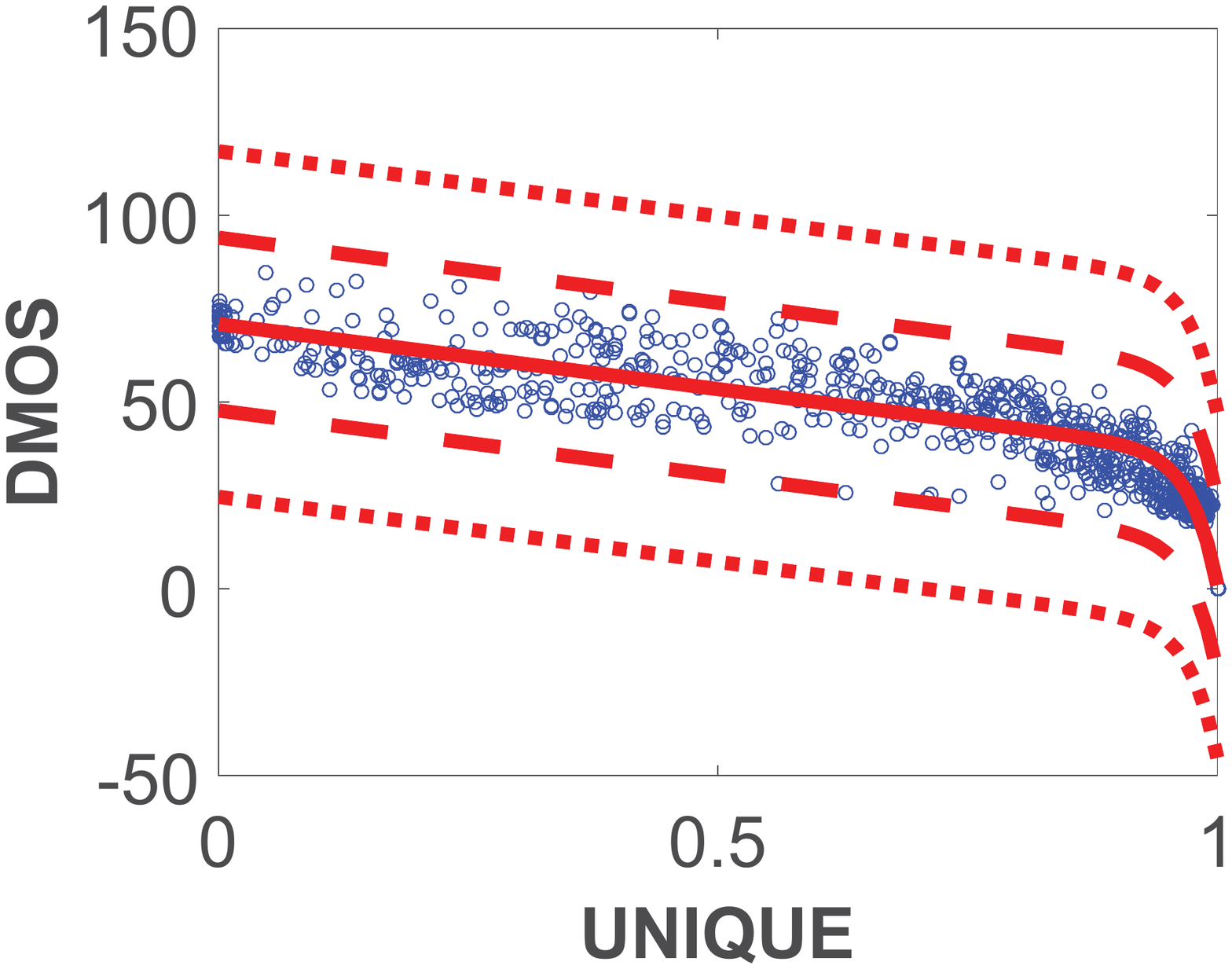}
  \vspace{0.03cm}
  \centerline{\footnotesize{(j)LIVE-UNIQUE}}
      \vspace{-0.45cm}
\end{minipage}
 \vspace{0.2cm}
\hfill
\begin{minipage}[b]{0.28\linewidth}
  \centering
\includegraphics[width=0.8\linewidth, trim= 20mm 75mm 20mm 65mm]{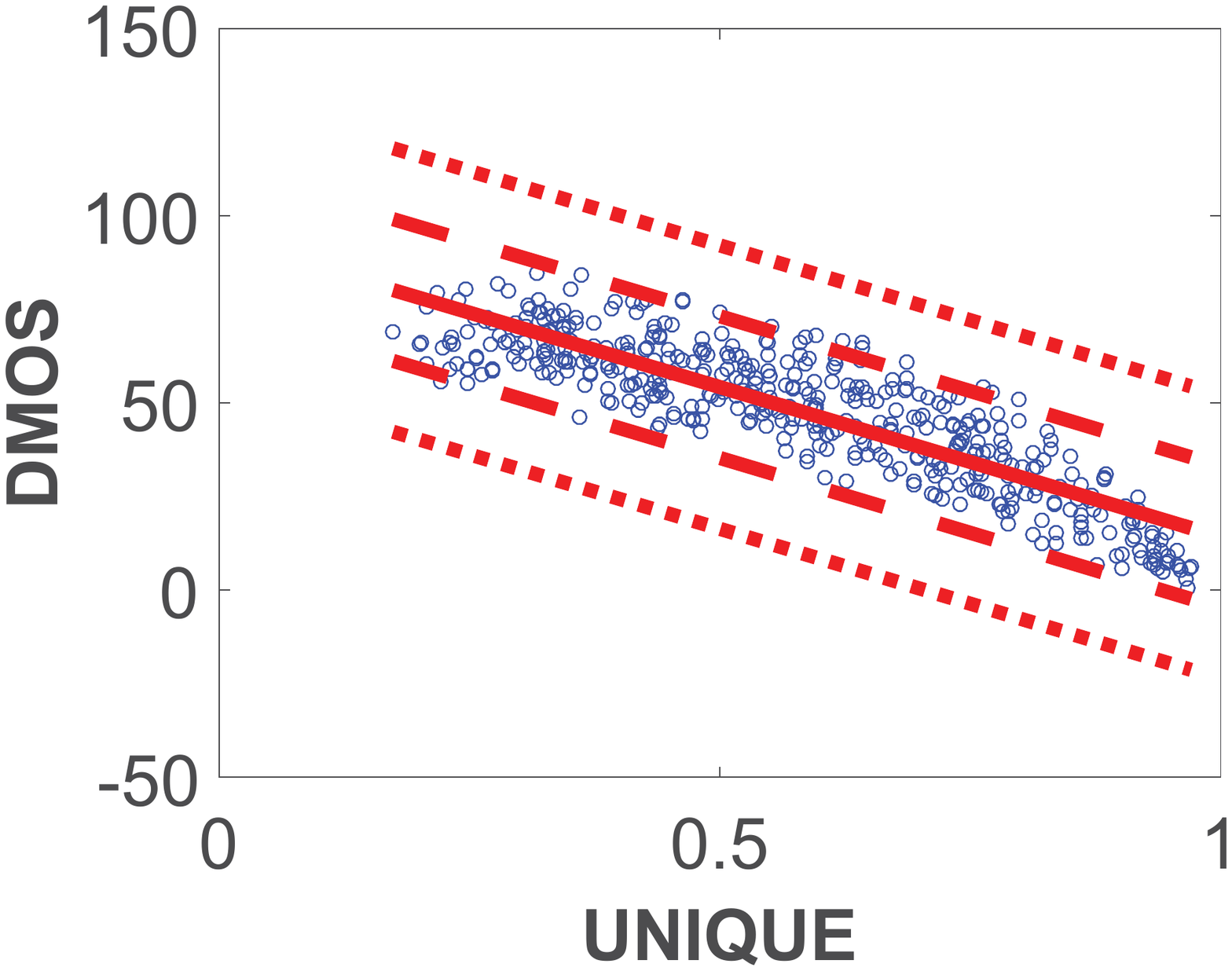}
  \vspace{0.03 cm}
  \centerline{\footnotesize{(k) MULTI-UNIQUE   } }
      \vspace{-0.45cm}
\end{minipage}
 \vspace{0.2cm}
\hfill
\begin{minipage}[b]{0.28\linewidth}
  \centering
\includegraphics[width=0.8\linewidth, trim= 20mm 75mm 20mm 65mm]{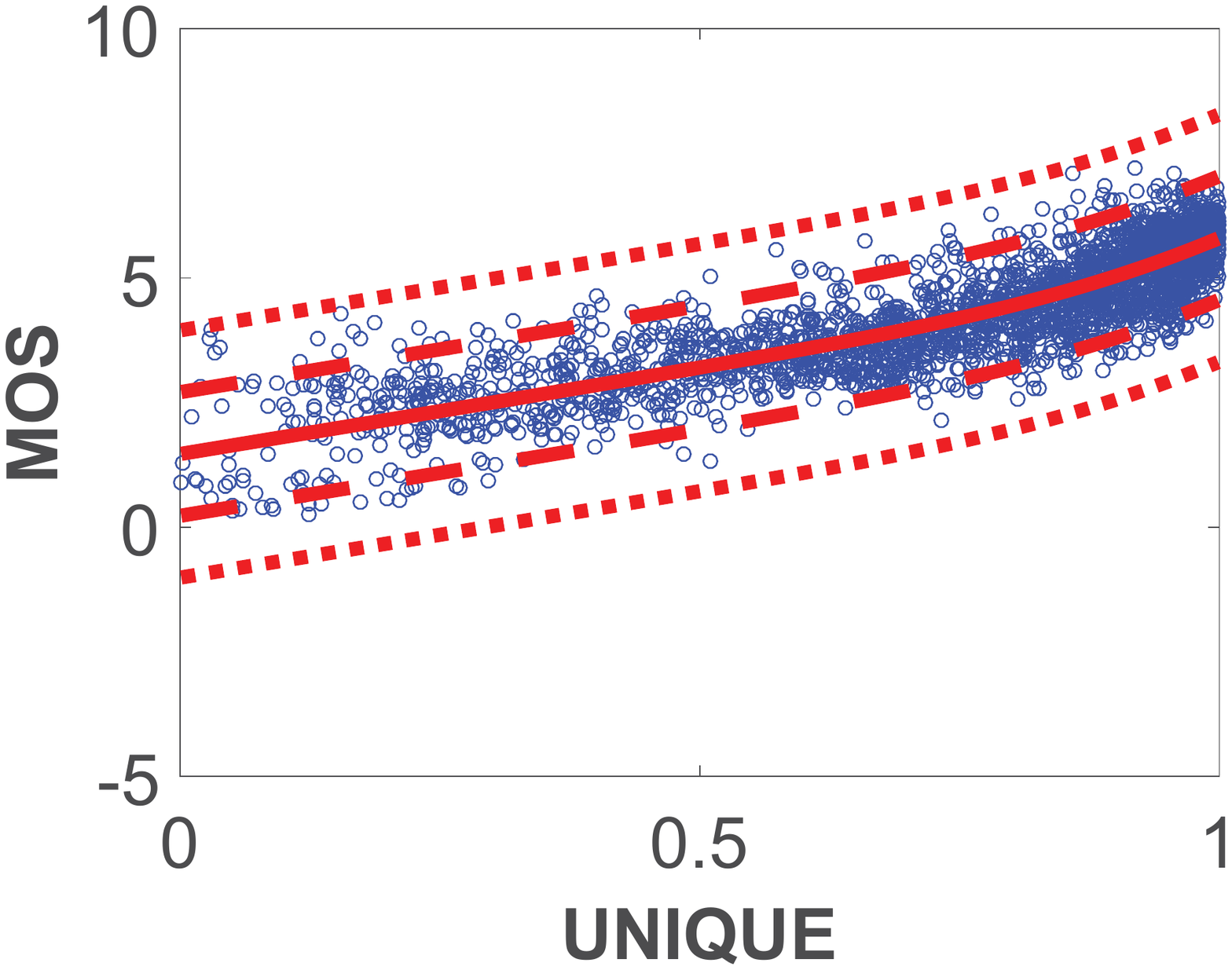}
  \vspace{0.03cm}
  \centerline{\footnotesize{(l) TID-UNIQUE }}
      \vspace{-0.45cm}
\end{minipage}

\caption{Scatter plots of outperforming quality estimators.}\vspace{-.5cm}
\label{fig:Scatter}
\vspace{-2.0mm}
\end{figure}

\end{center}

\vspace{-2.0mm}

To analyze the distribution of subjective scores versus objective quality estimates, scatter plots of the quality estimators that are among the two best performing methods in at least two different databases are given in Fig. \ref{fig:Scatter}, in which x axis corresponds to the quality estimates and y axis corresponds to the mean opinion scores (MOS) or the differential mean opinion scores (DMOS). For an ideal quality estimator, scores should be located on a linear curve. Therefore, in practice, we target scattered points that follow a linear pattern with low deviation. IW-SSIM, SR-SIM, and FSIMc are mostly located around high quality regions and they follow a monotonic pattern, whose decrease or increase is sharper close to the ideal quality score. On the contrary, \texttt{UNIQUE} scores are distributed all over the score range and follow a more linear behavior. Therefore, even without regression, \texttt{UNIQUE} is closer to an ideal estimator. We calculate the difference between the normalized histograms of ground truths and regressed objective quality estimates, whose scatter plots are reported, and highlight the methods that lead to a minimum difference in Table \ref{tab:hist_dist}. \texttt{UNIQUE} leads to the minimum difference in all databases and categories. Overall, the proposed method \texttt{UNIQUE} is consistently among the top performing quality estimators in $24$ out of $26$ categories whereas the closest best performing quality estimator is only in $7$ categories. More specifically, \texttt{UNIQUE} is the best performing metric in $19$ out of $26$ categories whereas the closest method is best performing in only $4$ categories.

\vspace{-2.0mm}

\section{Conclusion}
We proposed estimating perceived quality through monotonic relation between sparse representations of compared images. Preprocessing and postprocessing blocks are used to increase descriptiveness of chroma information, minimize redundancy in the spatial representations, and partially formulate suppression mechanisms in a visual system. The performance of \texttt{UNIQUE} shows that unsupervised learning-based sparse representations, which do not require distortion specific data or subjective opinions in the training, can robustly estimate perceived quality.


\end{document}